\documentclass[letterpaper]{article} % DO NOT CHANGE THIS
\usepackage{aaai24}  % DO NOT CHANGE THIS
\usepackage{times}  % DO NOT CHANGE THIS
\usepackage{helvet}  % DO NOT CHANGE THIS
\usepackage{courier}  % DO NOT CHANGE THIS
\usepackage[hyphens]{url}  % DO NOT CHANGE THIS
\usepackage{graphicx} % DO NOT CHANGE THIS
\usepackage{natbib}  % DO NOT CHANGE THIS AND DO NOT ADD ANY OPTIONS TO IT
\usepackage{caption} % DO NOT CHANGE THIS AND DO NOT ADD ANY OPTIONS TO IT
\usepackage[linesnumbered,ruled,vlined]{algorithm2e}
\usepackage{amsmath}
\usepackage{amsfonts}
\usepackage{csquotes}
\usepackage{multirow}
\usepackage{booktabs}
\usepackage{xcolor}
\title{Generative Adversarial Network with Soft-Dynamic Time Warping and Parallel Reconstruction for Energy Time Series Anomaly Detection}
\author {
    Hardik Prabhu \textsuperscript{\rm 1},
    Jayaraman Valadi \textsuperscript{\rm 2},
    Pandarasamy Arjunan \textsuperscript{\rm 3}
}
\affiliations {
    \textsuperscript{\rm 1, \rm 3}Robert Bosch Centre for Cyber Physical Systems, Indian Institute of Science, Bengaluru, India\\
    \textsuperscript{\rm 2}School of Computing and Data Sciences, FLAME University, Pune, India\\
    hardik.prabhu@gmail.com, jayaraman.vk@flame.edu.in, samy@iisc.ac.in
}

\usepackage{bibentry}
\begin{document}

\maketitle

%itle 1: Generative Adversarial Network with Parallel Reconstruction for Anomaly Detection in Building Energy Time Series

%Title 2: Energy Anomaly Detection using Generative Adversarial Network with Soft-Dynamic Time Warping

%Title 3:Generative Adversarial Network with Soft-Dynamic Time Warping and Parallel Reconstruction for Energy Time Series Anomaly Detection

\begin{abstract}
In this paper, we employ a 1D deep convolutional generative adversarial network (DCGAN) for sequential anomaly detection in energy time series data. % trained using Wasserstein loss on normal energy usage data.
Anomaly detection involves gradient descent to reconstruct energy sub-sequences, identifying the noise vector that closely generates them through the generator network. Soft-DTW is used as a differentiable alternative for the reconstruction loss and is found to be superior to Euclidean distance. Combining reconstruction loss and the latent space's prior probability distribution serves as the anomaly score. Our novel method accelerates detection by parallel computation of reconstruction of multiple points and shows promise in identifying anomalous energy consumption in buildings, as evidenced by performing experiments on hourly energy time series from 15 buildings. 
%real-world datasets.
%In this paper, we employ a 1D deep convolutional generative adversarial network (DCGAN) for sequential anomaly detection in energy time series data trained using Wasserstein loss on normal energy usage data. Anomaly detection involves gradient descent to reconstruct energy sub-sequences, identifying the noise vector that closely generates them through the generator network. Soft-DTW is used as a differentiable alternative for the reconstruction loss and is found to be superior to Euclidean distance. Combining reconstruction loss and the latent space's prior probability distribution serves as the anomaly score. Our novel method accelerates detection by parallel computation of reconstruction of multiple points and shows promise in identifying anomalous energy consumption in buildings, as evidenced by performing experiments on real-world datasets.
\end{abstract}

\section{Introduction}
Buildings consume significant energy, accounting for approximately 40\% of total energy usage worldwide. 
The recent proliferation of smart metering systems has led to an unprecedented volume of energy time-series data.
Through data-driven analysis, valuable insights about the buildings' energy use patterns have been obtained, offering informative perspectives on energy usage. However, it is crucial to acknowledge the influence of anomalies on energy management. In commercial buildings, the presence of inadequately maintained, faulty, or deteriorated hardware, along with improper operational practices, is estimated to contribute to the wastage of 15 to 30 \% of energy consumption \cite{katipamula2005methods,schein2006rule}. Furthermore, if anomalous energy use instances are not accurately identified and properly corrected, it can distort the reference points used for making predictions, leading to inaccurate and unreliable future forecasts as well. Hence, energy anomaly detection is essential for efficient energy management.  The existing methods in this field primarily focus on identifying power samples that show significant deviations from the normal consumption patterns, either being excessively high or low. However, this approach may not effectively capture the sequential anomalies that could indicate more complex issues in energy usage \cite{himeur2021artificial}.

In recent years, generative models have gained a lot of attention due to their remarkable ability to learn and mimic complex data distributions, leading to advancements in diverse fields such as image synthesis, natural language processing, and time series analysis. Generative models such as Generative Adversarial Networks (GANs) \cite{goodfellow2014generative} can effectively learn the patterns, trends, and dependencies present in normal data and then generate synthetic data points that closely resemble the original data. GANs are good at synthesis, particularly time series, and are widely used for anomaly detection \cite{di2019survey} despite the challenges faced in the training of GANs, such as vanishing gradients, non-convergence, and diminishing gradients \cite{brophy2021generative}. Their superiority in effectively identifying intricate features in time series data, aids in the detection of complex anomalies. This makes them stand out in comparison with the previously used anomaly detection models \cite{zhu2019novel}.

Unlike the statistical likelihood-based anomaly detection models such as those used in \cite{coluccia2013distribution}, GANs do not rely on identifying the likelihood of a data point coming from the dense region of the data distribution. Instead, GANs effectively learns a mapping between a simple prior, typically the standard Gaussian distribution in a latent space to the data distribution.  The mapping of the data distribution to the normal distribution is then followed by the reconstruction of data points from the normal distribution. If the GAN is unable to generate (reconstruct) a particular sample accurately, it can be logically concluded that the sample is an anomaly \cite{li2019mad,bashar2020tanogan,geiger2020tadgan}. In our work, we invert the generator network (inverse mapping)  of the GAN for reconstruction by applying gradient descent in the latent space \cite{schlegl2017unsupervised}, and subsequently perform anomaly detection.

The main contributions of this paper are:
\begin{itemize}
    \item Efficient utilization of Generative Adversarial Networks (GANs) for anomaly detection in univariate energy time series, derived from meter readings from a dataset of real buildings.
    \item Proposing Soft-DTW \cite{cuturi2017soft} as an alternative to Euclidean distance as a differentiable reconstruction loss.
    \item Parallel computation of reconstruction of multiple data points simultaneously, effectively mitigating the performance bottleneck.
    \item Adapting a method for effectively evaluating sequence-level anomaly detection with GANs for dataset annotated with pointwise anomaly labels.
\end{itemize}

%The rest of the paper is organized as follows. Background and motivation are described in \ref{sec:background}.

\section{Background and Motivation}
\label{sec:background}

It has been reported that more than 20\% of the total energy consumed within buildings is wasted due to various factors~\cite{roth2005energy}. Therefore, it is important to identify these energy wastage events to reduce buildings' operational costs. Various existing research has centred on data-driven anomaly detection methods that utilize smart metering data~\cite{himeur2021artificial}.
Despite these advancements, robust anomaly detection in real buildings remains a challenging task due to the complex nature of anomalies caused by various deviations in the operational patterns of the buildings. While there is increasing interest among them, their primary focus is on detecting instantaneous energy wastage (point anomaly). Whereas, the studies are limited in scope in detecting continuous energy wastage events (sequence anomaly) which is a challenging task ~\cite{himeur2021artificial}.

In recent years, GANs have gained increasing prominence in detecting complex anomalies across various domains. Their capability to capture complex patterns and dependencies in normal data enables them to generate synthetic data points that closely resemble the original data, making them effective tools for anomaly detection. While various variants of GANs are available, in this paper, we employ a variant known as the 1-dimensional Deep Convolutional Generative Adversarial Network (1D-DCGAN) \cite{radford2015unsupervised} for detecting anomalies in building energy time series data. %In the next sections, we present how GANs work and discuss their suitability for anomaly detection.

%We propose using a 1D Deep Convolutional Generative Adversarial Network (DCGAN) framework for unsupervised anomaly detection in such univariate time series data. Unlike some prior approaches which rely on recurrent neural networks for both the generators and the discriminator, 1D convolutional networks provide computational efficiency when modelling high-frequency time series. We train a 1D DCGAN on normal energy usage sub-sequences to learn a model which is capable of generating standard consumption patterns. To detect anomalies, we reconstruct a given energy sub-sequence by using gradient descent to find the noise vector that most closely generates the sub-sequence when passed through the generator network. The reconstruction error between the original sub-sequence and the reconstructed sub-sequence is used as an anomaly score. We evaluate this framework on real-world building energy time-series datasets. Our experiments aim to show that the reconstruction error correlates with the presence of anomalies, demonstrating the potential of this methodology for anomaly detection in energy consumption data.
\subsection{Generative Adversarial Networks (GANs)}
A GAN consists of two networks, a generator (G) and a discriminator (D) which train in an adversarial setting. The main idea behind GANs is to have the generator network learn to generate data that resembles samples from a target data distribution, without explicitly modelling the distribution itself. The generator takes in random noise as input and tries to produce synthetic data samples that mimic the real data. The discriminator is trained to distinguish between real data samples from the actual dataset and fake data samples produced by the generator.
The value function $V(G, D)$ for the min-max game played between the generator and the discriminator is given below.

\begin{equation}
\begin{aligned}
 \min_{G}\max_{D}V(G,D)= 
 & \mathbb{E}_{x\sim p_{\text{data}}(x)}[\log{D(x)}] +  \\
 & \mathbb{E}_{z\sim p_{\text{z}}(z)}[1 - \log{D(G(z))}]   
  \end{aligned}
\label{eq:minmax}
\end{equation}

With a prior distribution $p(z)$ in the noise space, the generator implicitly defines a distribution $p_g$ for the generated output $G(z)$. The discriminator assigns a probability score $D(x)$ to input $x$ for belonging to the target distribution $p_{data}$. 
%The goal of adversarial training is to have $p_g$ which is close to $p_{data}$. 
In the original GAN~\cite{goodfellow2014generative}, several theoretical guarantees are derived which are important to our ongoing discussion. For a fixed G, the optimal discriminator ($D^{*}$) is given as  $D^{*} = \max_{D} V(G, D)$. The optimal discriminator could be then written in terms of the generator distribution ($p_{g}$) and the data distribution ($p_{data}$).
\begin{equation}
\label{eq:Dstar}
D^{*}(x)= \frac{p_{data}(x)}{p_{data}(x) + p_{g}(x)}
\end{equation}
By substituting the optimal discriminator $D^{*}$ in $V$, which is implicitly a function of $G$, the equilibrium of the min-max game is obtained by minimizing $C(G) = V(G, D^{*})$. The simplified expression is given below.
\begin{equation}
\label{eq:JSDgan}
C(G) = -\log(4) + 2 \cdot \text{JSD}(p_{\text{data}} \| p_g)
\end{equation}

The Jensen–Shannon divergence (JSD) is non-negative and only reaches its minimum when $p_g = p_{data}$. Minimizing the JSD between the data distribution and the generator's output is desirable for data generation. However, this approach might be overly cautious, leading to mode collapse, where the generator produces limited and repetitive samples without exploring the full diversity of the data distribution.
To prevent false positives during anomaly detection, it becomes essential to learn how to generate from the entire distribution of normal data. Mode collapse and other stability issues have led to the evolution of alternative loss functions.
%to address the limitations of original adversarial loss. Wasserstein Generative Adversarial Network (WGAN) introduces an alternative loss function based on optimal transport theory. 

\subsection{GAN Training with Wasserstein Loss}

W-GAN \cite{arjovsky2017wasserstein} suggests an alternative training process to provide a more stable and reliable training process for GANs. Wasserstein distance measures the distance between two probability distributions. It effectively tackles issues like the vanishing gradient and mode collapse, making it a favourable choice for enhancing the training and performance of GAN models. The theoretical formulation of the distance between $p_g$ and $p_{data}$ is given below. 
\begin{equation}
     W(p_{data},p_g) = \inf_{\gamma \in \Gamma(p_g, p_{data})} \mathbb{E}_{(x, y) \sim \gamma} [\|x - y\|]
\end{equation}
Where $\gamma$ is the joint distribution over $(x,y)$ such that the marginal distributions of x and y are $p_{g}$ and $p_{data}$. A more tractable version of the loss is given by the Kantorovich-Rubinstein duality.
\begin{equation}
    W(p_{\text{data}}, p_g) = \sup_{D \in \text{Lip}(K)} \left( \mathbb{E}_{x \sim p_{\text{data}}}[D(x)] - \mathbb{E}_{y \sim p_g}[D(y)] \right)
\end{equation}

A function (discriminator) \(D\) is \(K\)-Lipschitz if for all \(x\) and \(y\) in the domain of \(D\), the following condition holds:

\begin{equation}
    |D(x) - D(y)| \leq K \cdot \|x - y\|
\end{equation}

Therefore, the generator and discriminator are required to optimize the following min-max function.
\begin{equation}
    V(G,D) = \min_{G} \max_{D \in \text{Lip}(K)} \left( \mathbb{E}_{x \sim p_{\text{data}}}[D(x)] - \mathbb{E}_{y \sim p_g}[D(y)] \right)
\end{equation}

It is important to note that the discriminator is no longer limited to being a classifier; it can be any function as long as it adheres to the Lipschitz constraint. To maintain this constraint during training, gradient clipping is applied (See~\cite{arjovsky2017wasserstein}). Additionally, the loss function of the discriminator could be used to evaluate convergence. The absolute value of the discriminator loss in WGAN serves as an approximation of the Wasserstein distance between $p_{data}$ and $p_g$.

% \begin{figure}[t!]
% \centering
% %\includegraphics[width=0.36\linewidth,height =0.22\linewidth]{gan.drawio.png} 
% %\includegraphics[scale=0.7]{gan.drawio.png} 
% \includegraphics[scale=1.1]{gan-reconstucion-error.drawio.pdf}
% \caption{Reconstruction error obtained through GAN}
% \label{fig:reconstruction-error}
% \end{figure}

\subsection{Anomaly Detection Using GAN Inversion}
The GAN in our study is trained solely on normal data, which means its generator is designed to produce only realistic samples. If the generator is unable to replicate a certain sample, it can be inferred that the sample is an anomaly.
GANs, being likelihood-free models that generate realistic samples, require a mechanism to invert this generation process. This inversion maps the data back to the corresponding noise in the latent space, which is then passed through the generator. Comparing this output with the original data point helps in anomaly detection. 

While TadGAN \cite{geiger2020tadgan} uses training an encoder along with the generator for inverse mapping, we have elected to implement a different strategy. Our approach focuses on gradient descent in the latent space,  which involves separate processes for generation and reconstruction. This distinction enables us to use any differentiable loss, providing us with more flexibility in how we reconstruct data. Moreover, this way the success of our reconstruction process is largely dependent on the quality of the generator and the choice of the differentiable reconstruction error. 

Drawing inspiration from AnoGAN \cite{schlegl2017unsupervised}, which uses a Deep Convolutional GAN (DCGAN) trained on normal image datasets, we also use DCGAN, in a one-dimensional version suited for time series data.  To address the stability challenges in GANs highlighted by \cite{arjovsky2017wasserstein}, we implement W-GAN for training our model. 

Bashar et al. \cite{bashar2020tanogan} performed inverse mapping for time-series anomaly detection using gradient descent in latent space. However, they have encountered the same limitations as AnoGAN, it requires performing extensive optimization steps for reconstruction of each data point, which results in poor test-time performance \cite{di2019survey}. We address this challenge by utilizing parallel computation to simultaneously reconstruct multiple points. The details are present in the methodology section.

\section{Methodology}
Given an univariate time series $\{x_t\} = \{x_{1}, x_{2}, \ldots  x_{T}\}$, where $x_{i} \in R$ is a
meter reading at time step $i$. The objective is to devise a methodology to find a subset $A \subset \{x_t\} $ such that it contains the points that deviate from the normal energy usage pattern. We define the following terms:
%In this section, we will delve into the implementation details of our chosen methodology. However, before going into the specifics, we would like to introduce specific terminology to avoid any possible confusion.

\begin{enumerate}
%\item Time series : A complete set of chronological observations $X = \{x_{1}, x_{2}, .., x_{T}\}$  
\item Time segment: A subset of consecutive points from a time series. For example $S = \{x_{4}, x_{5}, x_{6}, \ldots x_{103}\}$.
\item Time sub-sequence: It is a smaller segment of the time series consisting of a fixed length, given by the variable window size.

\end{enumerate}
%\bluecolor{TODO: the above para should be revised.}

\subsection{The LEAD Dataset and Pre-processing}

The LEAD1.0 dataset \cite{gulati2022lead1} is used in this study. This public dataset includes hourly-based electricity meter readings for commercial buildings over up to one year. Each building contains about 8,784 data points. Anomaly annotations are provided, marking individual anomalous points within each building's time series. In this study, we selected 15 buildings with adequate normal data to assess the performance of the proposed framework.
 
\subsubsection{Train-test Segments}
First, the time series data for individual buildings' electricity consumption is partitioned into segments. Subsequently, a fixed-size rolling window is applied to each segment, creating time sub-sequences that serve as input to the model. For each of the annual meter reading time series in our dataset, we first remove missing readings. Then we divide each series into 25 contiguous, non-overlapping time segments. The segments contain point-wise anomaly annotations. The segments which have no anomalous points are used in training. The rest of the segments are used for testing. Each segment is normalized to be in the range [-1,1].  This step is crucial because our generator model utilizes a \textit{tanh} activation function at its end, which outputs within this specific range.

\subsubsection{Model Input}
A segment of time series data is further processed using a rolling window of fixed length to generate the model inputs. Let $S = \{x_{1},x_{2}, \ldots x_{n+w-1}\}$ be a segment of time series. Then, using window size $w$, a collection of $n$ time sub-sequences $\textbf{X}$ is generated as shown below.
\begin{equation}
    \textbf{X} = \{ (x_1, \ldots x_{w}) , (x_2, \ldots x_{w+1}),  \ldots (x_n, \ldots x_{n+w-1}) \}
\end{equation}  

The sub-sequences are overlapping in nature. The sub-sequences undergo further processing to form the model input, resulting in a tensor with the shape \textit{(Batch size, features (1), window size (w))}.

\subsection{1-D DCGAN Architecture}
% \begin{figure}[ht!]
% \centering
% \includegraphics[scale=0.6]{gan.png} 
% \caption{1D DCGAN architecture}
% \label{fig:dcgan}
% \end{figure}

\begin{figure}[t!]
\centering
\includegraphics[scale=0.64]{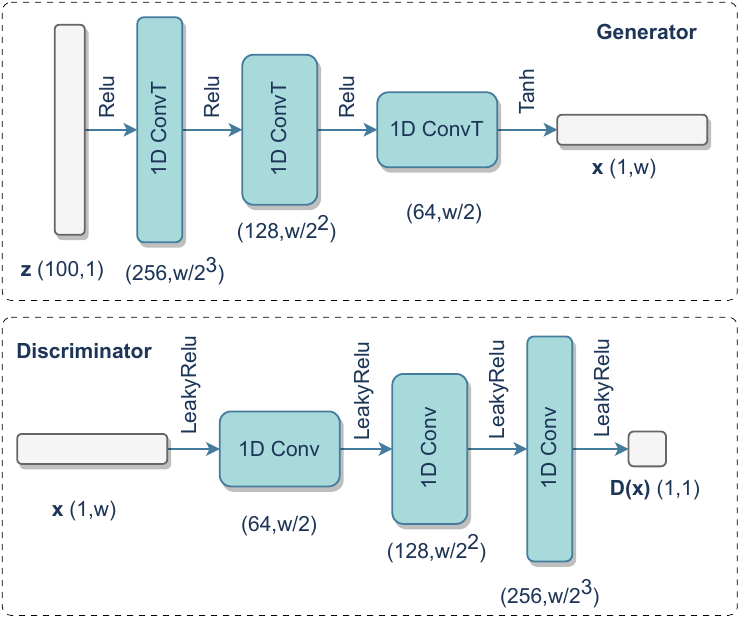}
 
\caption{1D DCGAN architecture for anomaly detection.}
  
\label{fig:dcgan}
\end{figure}

% \begin{figure*}[ht!]
% \centering
% \includegraphics[scale=0.15]{gan1.png} 
% \caption{1D DCGAN architecture}
% \label{fig:dcgan}
% \end{figure*}

Both the generator and discriminator have 
3 symmetrical hidden layers as shown in Figure \ref{fig:dcgan}. The generator uses a series of convolutional transpose layers to upsample the input noise vector into a 1D time series output. starting with a 1D transposed convolutional layer with 100 input channels (noise dimensions) and 256 output channels, followed by \textit{ReLU} activation, and three more transposed convolutional layers with decreasing output channels (128, 64, and 1), followed by \textit{Tanh} activation at the end. The discriminator consists of 4 1D convolutional layers, each followed by  \textit{Leaky ReLU} activation. In a DCGAN, batch normalization is commonly used in intermediate layers to stabilize training and improve convergence. However, it is typically not used in the generator's output layer and in the discriminator's input layer and output layer. %We make the window size (w) of the input sub-sequence to be of any value as long as it is divisible by 8. 
The latent space is made up of 100 dimensions and we take the prior distribution $p_z$ as an independent standard Multivariate Gaussian over the latent space.

\subsection{Inverse Mapping}

The goal of inverse mapping is to estimate the input that generated a particular output. In the context of deep learning, it is the reversing of the forward mapping that a neural network might have learned. Inverse mapping is an active area of research in representation learning, generative modelling, explainability and adversarial robustness. Key technical approaches include encoder-decoder architectures, conditional generation, and inverting using gradient descent in the input space. 

In this work, we focus on the gradient descent approach in the latent (input) space. The idea is straightforward. To invert the query data $X$. It requires starting from noise in latent space $Z \sim p_z$ and updating it using gradient descent by evaluating a differentiable (reconstruction) loss $L(X, G(Z))$. After few iterations, $Z$ is considered as $G^{-1}(X)$. The details are described in Algorithm 1.

\begin{algorithm}[t!]
\label{algo: inverse_map}
\caption{Inverse Mapping with Gradient Descent}
\SetKwInput{Requires}{Requires}
\Requires{Query sub-sequence $X$, Generator $G$, Latent space prior $p_z$, Learning rate $\alpha$, Reconstruction loss $L$}
\KwResult{Reconstructed sub-sequence $X'$}
Initialize random latent vectors $Z \sim p_z$ ;
\While{stopping criteria not met}{
    Generate synthetic data samples using the generator: $X_{r} = G(Z)$\;
    Calculate the reconstruction error: $Loss = L(X, X_{r})$\;
    Calculate the gradient of the loss w.r.t. $Z$: $\nabla_{Z}L = \frac{\partial Loss}{\partial Z}$\;
    Update the latent vector using gradient descent: $Z = Z - \alpha \cdot \nabla_{Z}L$\;
}
Reconstruct data sub-sequence using the updated latent vector: $X' = G(Z)$\;
\end{algorithm}
   
\subsubsection{Reconstruction Loss}
 When evaluating the error (dissimilarity) between two time series sub-sequences, Dynamic Time Warping (DTW) \cite{berndt1994using} is often used as a loss function rather than mean squared error (MSE). DTW allows for flexible alignment between the original and reconstructed time sub-sequence, overcoming small shifts and distortions. MSE assumes a fixed one-to-one alignment and penalizes any mismatch equally, even if just slightly misaligned. %DTW identifies the minimum cost alignment through dynamic programming. 
DTW compares the overall shapes of the time sub-sequences, rather than exact value matching. This is more appropriate for generative modelling where we need to capture the essence of temporal patterns and dynamics, not necessarily reproduce the original numerical values. We use Soft-DTW  \cite{cuturi2017soft} as a differentiable alternative to DTW for reconstruction loss, enabling gradient descent-based optimization. 

Given two sequences $X = (x_1, x_2, ..., x_n)$ and $Y = (y_1, y_2, ..., y_m)$, the Soft-DTW distance can be defined using the following recurrence relation:

\begin{equation}
\begin{aligned}
    D[i,j] = & \text{softmin}_\gamma \left( D[i-1,j], D[i,j-1], D[i-1,j-1] \right) \\
    & + d(x_i, y_j) 
\end{aligned}
\end{equation}

where $d(x_i, y_j)$ is a distance metric between the elements $x_i$ and $y_j$, and $\text{softmin}_\gamma$ is the soft minimum operator \cite{cuturi2017soft}. The Soft-DTW distance between the two sequences 
X and 
Y is found at 
$D[n,m]$.

\subsubsection{Parallel Computation} The way the neural networks are designed, the generator $G$ can perform parallel computations to construct $k$ number of data points in data space from $k$ noise vectors.
These could further compared with $k$ query points $\textbf{X} = [X_1,X_2, ... X_k]$  to get a loss vector. 

\begin{equation}
    L(G(\mathbf{Z}),\mathbf{X}) = \begin{bmatrix}
    L(G({Z}_1),X_1) \\
    L(G({Z}_2),X_2) \\
    L(G({Z}_3), X_3) \\
    \vdots  \\
    L(G({Z}_k), X_k)  
\end{bmatrix}
\end{equation}

If the final loss in the backward computation is the mean (or the sum) of all the losses, then since the weights of the network are not being updated, all the points could be updated simultaneously.  

\begin{equation}
    \textbf{Z} = \textbf{Z} - \alpha \nabla_\textbf{Z} \frac{1}{k}\sum_{i=1}^{k}(L(G(Z_i),X_i) )
\end{equation}

Hence, the reconstruction operations can be executed in parallel for multiple data points. When combined with GPU acceleration, our experiments demonstrate a significant reduction in computation time. Refer to Figure \ref{fig: reconst} for an illustration.

\subsection{Anomaly Score}
After training the GANs, any input sub-sequence \(X\) with the shape \((1, 1, w)\) can be inverted as \(Z\) and reconstructed as \(G(Z)\) of the same shape using Algorithm 1. Multiple sub-sequences could be reconstructed in parallel. The reconstruction loss is measured using Soft-DTW, which also provides a direct way of obtaining an anomaly score. Our objective goes beyond identifying if there is an appropriate input to generate a sample resembling \(X\). We also want to ensure that this input can be generated by sampling from the latent space distribution \(p_z\) which is Gaussian-centred at the origin. The combined anomaly score is presented below.
\begin{equation}
   \text{Anomaly score}(X) = \alpha*\text{Soft-DTW}(X,G(Z)) + \beta*\|Z \|_2 
\end{equation}

Where $\alpha$ and $\beta$ regulate the influence of each subpart. The input sub-sequence is marked as anomalous if the anomaly score exceeds a certain threshold.

\subsection{Anomaly Detection Using Kernel Density Estimates}
Due to the sequential nature of anomalies, a direct comparison between the ground truth and the prediction becomes difficult. In this regard, we use a similar method proposed in the prior work done by Gu et al. \cite{gu2022degan}. The set of time windows (sub-sequences) identified as an anomaly by the model are to be first converted into anomalous timestamps to make a comparison with the ground truth. This is achieved by performing the following steps.

\begin{enumerate}
\item Select a test segment for evaluation.
\item Create overlapping sub-sequences of window size w.
\item Produce anomaly scores for each of the sub-sequences.
\item For sub-sequences with anomaly scores greater than a set threshold, mark the timestamp corresponding to the middle of each sub-sequence as the critical points. 
\item Use Kernel Density Estimation (KDE) \cite{parzen1962estimation} to create a distribution over the test segment of the critical points. Scale the density to lie between 0 and 1. 
\item Find points of the scaled KDE above a certain height and mark those timestamps as the predicted anomalies.
\end{enumerate}

\begin{figure*}[t!]
\centering
\includegraphics[width=\linewidth,height=0.2\linewidth]{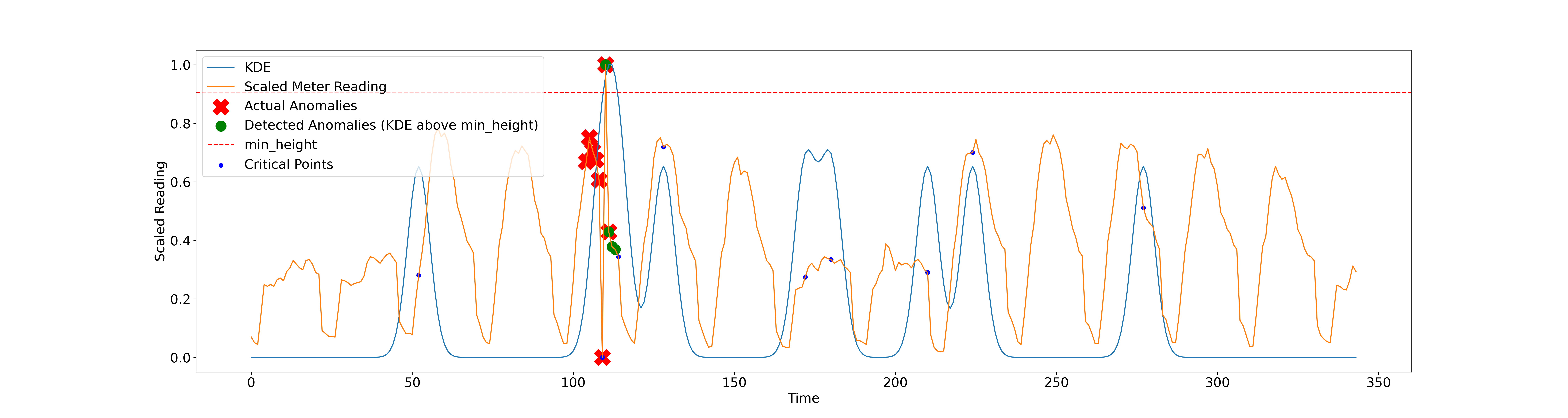} 

\caption{\small A test time series segment from a building with timestamps indexed starting from 0 along the x-axis. The orange line represents the scaled actual meter reading, and the blue line represents the scaled KDE. Actual annotated anomalies are indicated by red crosses, while our method marks anomalies with green dots, considering only KDE values above a threshold.}

\label{fig: Visual proof of concept}
\end{figure*}

Refer to Figure \ref{fig: Visual proof of concept}
for an illustration of anomaly detection over a test segment.

\subsection{Model Evaluation}
Although the labels are pointwise, time series anomalies can span a range of time, making it challenging to precisely define the onset of an anomaly. Thus, as long as the model predicts in the vicinity of the anomaly label, it should not be penalized. Introducing some tolerance ($r_t$) in performance calculations accounts for this consideration. To evaluate the model, we adopt precision, recall and F1 score. True Positives (TP), False Negatives (FN), and False Positives (FP) calculations are adjusted with the introduction of tolerance. 

\begin{equation}
\begin{aligned}
&TP: |d - p_{\text{closest}}| \leq r_t, \hspace{0.1cm} \\
&FN: |d - p_{\text{closest}}| > r_t, \hspace{0.1cm} \\
&FP: |p - d_{\text{closest}}| > r_t
\end{aligned}
\end{equation}

Where \(d\) represents the location of a labelled ground truth anomaly, and \(p_{\text{closest}}\) denotes the nearest anomaly prediction. Ground truth anomalies with a predicted anomaly in their proximity are considered TP, while those without any predicted anomaly nearby are classified as FN. FP refers to an anomaly prediction that does not correspond to any real defect in its vicinity, with \(p\) representing the location of the predicted anomaly and \(d_{\text{closest}}\) representing the nearest actual annotated anomaly location.

\section{Experiments and Results}
The implementation of 1D-DCGAN is done using PyTorch, a popular deep-learning library in Python. 
In our experimental setup, we use the Adam optimizer for training both networks, with the beta set to 0.5 and a learning rate of 0.0002. The WGAN's "ncritic" parameter is configured to 5, and the clipping value is set at 0.01. Additionally, a batch size of 128 is used. Our analysis focuses on sub-sequences with a window size of 48 and a latent space of 100 dimensions. The Gan model is trained for 200 epochs. The evaluation hyperparameters for anomaly scores are adjusted while observing the model performance on the test set. 

\begin{figure}[ht!]
\centering
\includegraphics[width=\linewidth]{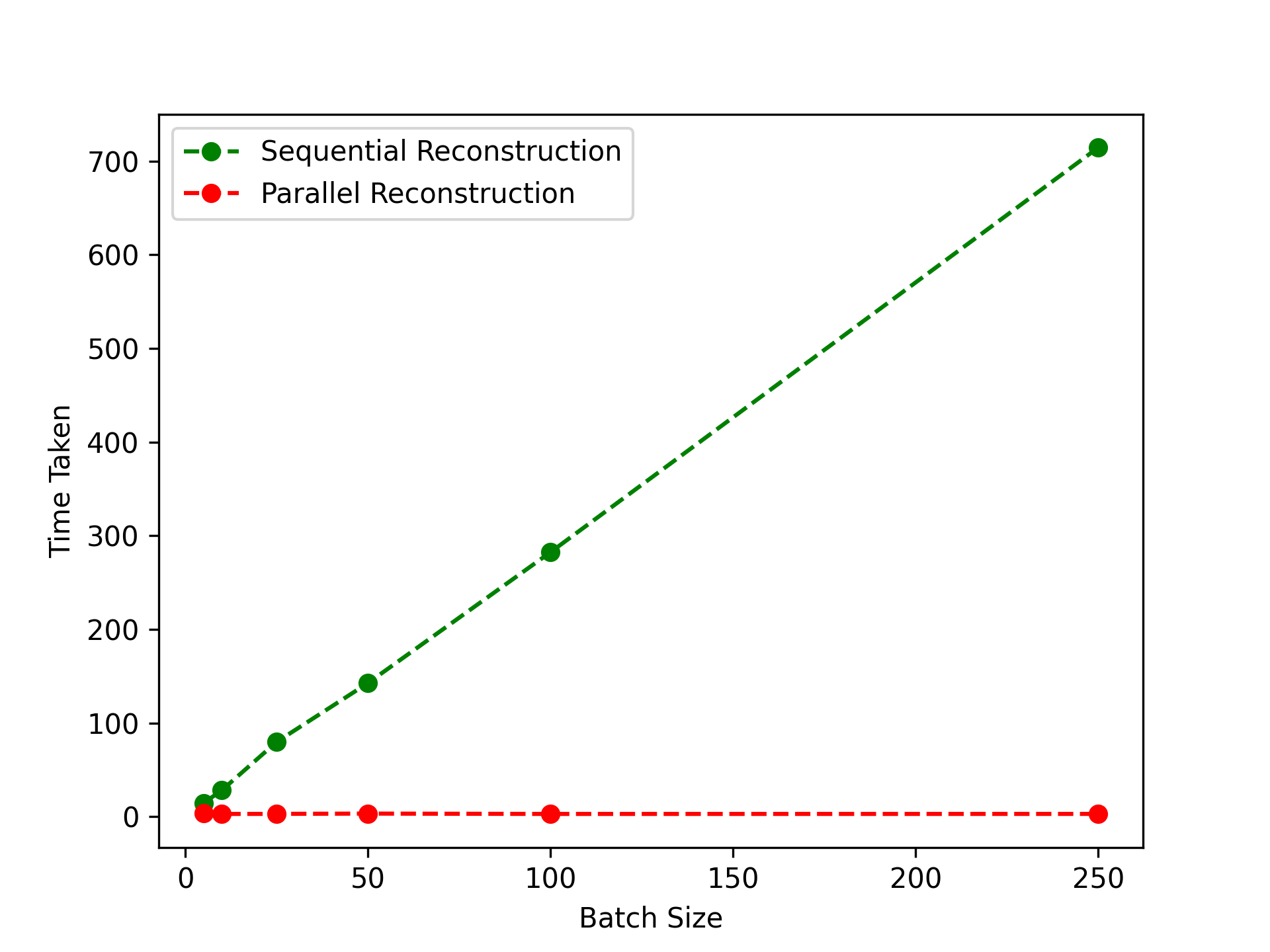} 

\caption{Comparison between time taken for reconstruction done using Mean Square Error loss sequentially and in parallel for different batch sizes.}

\label{fig: reconst}
\end{figure}

We evaluate our methodology on each energy time series separately.
The time series for each building is divided into train-test segments. The sub-sequences are created using a window size of 48 hours.  The training segments are distinct from the test segments. For the evaluation, we set the hyperparameters separately for each building: 1. \( \alpha \), 2. \( \beta \), 3. anomaly score 'threshold', and 4. KDE 'min\_height'. Figure \ref{fig: Visual proof of concept} demonstrates the anomaly detection process over a test time segment. The x-axis shows timestamps in hours indexed from zero. The y-axis contains the scaled meter readings (orange line). After tuning the evaluation hyper-parameters for a building, a test segment for that building is evaluated. The critical points are first calculated (mid-point of the anomalous subsequences), which are marked as blue dots. Then a kernel density (KDE) is fitted over the critical points (line in blue). The points which have a scaled KDE value above the min\_height (dotted red horizontal line) are marked as anomalous (green dots).The red crosses indicate the actual annotated anomalies. Since the actual anomalies are annotated by human inspection, and it is difficult to pinpoint the exact onset of a sequential anomaly, it is important that our model marks the regions where the sequential anomaly took place rather than marking the exact annotated anomaly points. Therefore, some tolerance is incorporated when calculating the F1 score.
Table \ref{tab:performance-metrics} compares the performance metrics of the anomaly detection model for different tolerance ($r_t$) values.

Reconstruction done in parallel significantly decreases the testing time. Figure \ref{fig: reconst} demonstrates the time (measured in seconds) required to reconstruct data across different batch sizes on the same machine. When reconstructing one data point at a time (sequential reconstruction), the time taken displays a linear relationship with the batch size. In contrast, when reconstruction is performed in parallel, the time taken remains relatively constant across different batch sizes.

\begin{table}[t!]  % Use 'table' for single column width
\centering
\caption{Performance metrics on different buildings}
\label{tab:performance-metrics}
\resizebox{\columnwidth}{!}{% Resize to fit within a single column width
\begin{tabular}{|l|rrr|rrr|}
\hline
\multirow{2}{*}{Building No.} & \multicolumn{3}{c|}{$r_t=12$} & \multicolumn{3}{c|}{$r_t=24$} \\
\cline{2-7}
& Rec. & Prec. & F1 & Rec. & Prec. & F1 \\
\hline
1 & 0.983 & 0.988 & 0.986 & 0.988 & 0.994 & 0.991 \\
2 & 0.629 & 0.991 & 0.769 & 0.697 & 0.992 & 0.819 \\
3 & 0.768 & 0.663 & 0.712 & 0.898 & 0.792 & 0.842 \\
4 & 0.668 & 0.725 & 0.695 & 0.747 & 0.827 & 0.785 \\
5 & 0.936 & 0.876 & 0.905 & 0.936 & 0.978 & 0.956 \\
6 & 0.618 & 0.870 & 0.723 & 0.751 & 0.903 & 0.820 \\
7 & 0.882 & 0.918 & 0.900 & 1.000 & 1.000 & 1.000 \\
8 & 0.674 & 0.796 & 0.730 & 0.838 & 0.856 & 0.847 \\
9 & 0.791 & 0.772 & 0.781 & 0.917 & 0.890 & 0.903 \\
10 & 0.692 & 0.759 & 0.724 & 0.835 & 0.792 & 0.813 \\
11 & 0.570 & 0.833 & 0.677 & 0.640 & 0.936 & 0.760 \\
12 & 0.627 & 0.941 & 0.753 & 0.637 & 1.000 & 0.778 \\
13 & 0.841 & 0.891 & 0.865 & 0.869 & 0.959 & 0.912 \\
14 & 0.945 & 0.937 & 0.941 & 0.945 & 1.000 & 0.972 \\
15 & 0.817 & 0.810 & 0.813 & 0.846 & 0.815 & 0.830 \\
\hline
Ave   & 0.763  & 0.851 & 0.798 & 0.836 &  0.915 & 0.869 \\
\hline
\end{tabular}
}
\end{table}

We conduct a comparative analysis of the average performance across all buildings using different reconstruction losses. Commonly, Euclidean distance is employed. However, our study confirms the effectiveness of Soft-DTW as a suitable alternative differentiable reconstruction loss. Since we are reconstructing multiple data points in parallel in a batch, our model operates in two modes: Active Statistics Mode (ASM) and Static Statistics Mode (SSM). In the ASM mode, the generator model's BatchNorm layers utilize the mean and variance of the current batch while performing reconstruction.
In contrast, in the SSM mode, they rely on the estimated population statistics gathered during training. This distinction can significantly impact the model's performance. We test the model using both reconstruction losses, and the results are detailed in Table \ref{tab:average-performance-metrics}. Notably, Soft-DTW proves to be the more effective method in both scenarios, achieving F1 scores of 0.869 and 0.834. This performance is notably better than that of the Euclidean loss, which achieves  F1 scores of 0.824 and 0.548. Interestingly, turning off the evaluation mode enhances the results specifically when using Soft-DTW.

\begin{table}
\centering
\caption{Average Performance Metrics for different choices of reconstruction loss}
\label{tab:average-performance-metrics}
\resizebox{\columnwidth}{!}{% Resize to fit within a single column width
\begin{tabular}{|c|ccc|ccc|}
\hline
\multirow{2}{*}{Reconstruction} & \multicolumn{3}{c|}{$r_t=12$} & \multicolumn{3}{c|}{$r_t=24$} \\
\cline{2-7}
                   & Rec. & Prec. & F1  & Rec. & Prec. & F1  \\
\hline
Soft-DTW (ASM mode) & 0.763  & 0.851     & 0.798    & 0.836  & 0.915     & 0.869    \\
Soft-DTW (SSM mode) & 0.785  &   0.832     & 0.777    & 0.843 &  0.879      &  0.834    \\
Euclidean (SSM mode) &  0.773  & 0.789     &  0.751    &  0.851  &  0.853    & 0.824    \\
Euclidean (ASM mode) & 0.516  & 0.520     & 0.476    & 0.603  & 0.582     & 0.548    \\

\hline
\end{tabular}
}
\end{table}

%\subsection{Anomaly detection on a set of 10 buildings}
%We also evaluated the model on a large set 10 buildings from LEAD1.0.

In this study, we also benchmark our model against an autoencoder. Specifically, we employ a 1-D Convolutional Neural Network (CNN) Autoencoder, which is designed with an encoder that mirrors the generator component of our Generative Adversarial Network (GAN), and a decoder that serves as its symmetrical counterpart. Additionally, we evaluate the performance of the Local Outlier Factor (LOF) model. All the models are evaluated on each of the buildings separately and then the average performance metrics are calculated and presented in Table \ref{tab:24hr-performance}. Our GAN-based model outperforms all others, particularly when employing Soft-DTW as the loss function, achieving an F1 score of 0.869 with a 24-hour tolerance. The 1D-CNN autoencoder ranks as the second-best model, securing an F1 score of 0.829, closely followed by the GAN model with Euclidean reconstruction, which scores 0.824 in F1. In contrast, the non-deep learning-based method, Local Outlier Factor (LOF), shows a significantly lower performance, with an F1 score of 0.414, indicating its ineffectiveness in detecting sequential anomalies.

\begin{table}
    \centering
    \caption{Average performance metrics across 15 buildings}
    \label{tab:24hr-performance}
    \begin{tabular}{ccccccc}
    \toprule
    Model & \multicolumn{3}{c}{$r_t =24$  (Tolerance)} \\
    \cmidrule(r){2-4}
    & Precision & Recall & F1 \\
    \midrule
    GAN (Soft-DTW)*   & 0.915  & 0.836 & 0.869   \\
    GAN (Euclidean)   & 0.853 & 0.851 & 0.824   \\
    1-D CNN-Autoencoder & 0.887
 & 0.840
 & 0.829
 \\
    LOF          &  0.305   &  0.815     & 0.414      \\
    \bottomrule
    \end{tabular}
    \\[1ex]
    \textit{*Best performing model with Soft-DTW reconstruction}
    
\end{table}

The source code and additional materials related to this study are available in the GitHub repository. \footnote{ \url{https://github.com/HardikPrabhu/Energy-Time-series-anomaly-detection}}

\section{Conclusion}

The paper systematically demonstrates how to train the GAN on normal data, perform gradient-based inverse mapping to reconstruct query samples, and use the reconstruction error as an anomaly score to generate critical points on a given time series segment and create an anomaly distribution on the segment using KDE. Tolerance is introduced for comparison with models that detect anomalies at the sub-sequence or pointwise level. This makes the evaluation more flexible. 

By incorporating parallel computation, we effectively mitigate the performance bottleneck, a commonly cited drawback of this approach. This enhancement opens avenues for more studies on various components in an isolated context. For example, refining the generative process through improved model architecture, optimizing the choice of reconstruction loss, and enhancing the anomaly scoring mechanism. Additionally, there is potential for exploring different approaches to benchmark sequential anomalies, which in most of the datasets, are annotated solely by timestamps (pointwise).

One limitation of our method, which involves using overlapping sub-sequences to identify critical points for a time segment, is the possibility of lower Kernel Density Estimation (KDE) values for the beginning and end parts. This reduction is attributed to the overlapping method used in deriving sub-sequences, leading to a lesser relative frequency of critical points in these specific areas. To address this issue, we could add a window length's worth of data at both the start and end of a test segment, borrowing from the preceding and subsequent segments, respectively.

While the results showcased in this study are preliminary, they underscore the potential and promise of the approach. However, it is crucial to recognize the need for more rigorous benchmarking. This entails comprehensive comparisons across a broader range of buildings and an extensive variety of models. Such thorough evaluations, which extend beyond the scope of our current work, will provide deeper insights and reinforce the findings presented in our paper. The limitations identified will be addressed in an expanded version of this paper.

\bibliography{aaai24}

\end{document}